\documentclass[lettersize,journal]{IEEEtran}
\usepackage{amsmath,amsfonts}
\usepackage{algorithmic}
\usepackage{algorithm}
\usepackage{array}
\usepackage[caption=false,font=normalsize,labelfont=sf,textfont=sf]{subfig}
\usepackage{textcomp}
\usepackage{stfloats}
\usepackage{url}
\usepackage{verbatim}
\usepackage{graphicx}
\usepackage{cite}
\usepackage{amssymb}
\begin{document}

\title{Explainable Human Activity Recognition: A Unified Review of Concepts and Mechanisms}

\author{Mainak Kundu, Catherine Chen, Rifatul Islam, Ismail Uysal, and Ria Kanjilal
\thanks{Mainak Kundu, Rifatul Islam, and Ismail Uysal are with the Department of Electrical Engineering, University of South Florida, Tampa, FL 33620 USA (e-mail: mkundu@usf.edu).}
\thanks{Catherine Chen and Ria Kanjilal are with the Department of Computer Engineering, California
Polytechnic State University, San Luis Obispo, CA 93407 USA (e-mail: rkanjila@calpoly.edu)}}

\maketitle

\begin{abstract}

Human activity recognition (HAR) has become a key component of intelligent systems for healthcare monitoring, assistive living, smart environments, and human–computer interaction. Although deep learning has substantially improved HAR performance on multivariate sensor data, the resulting models often remain opaque, limiting trust, reliability, and real-world deployment. Explainable artificial intelligence (XAI) has therefore emerged as a critical direction for making HAR systems more transparent and human-centered. This paper presents a comprehensive review of explainable HAR methods across wearable, ambient, physiological, and multimodal sensing settings. We introduce a unified perspective that separates conceptual dimensions of explainability from algorithmic explanation mechanisms, reducing ambiguities in prior surveys. Building on this distinction, we present a mechanism-centric taxonomy of XAI-HAR methods covering major explanation paradigms. The review examines how these methods address the temporal, multimodal, and semantic complexities of HAR, and summarize their interpretability objectives, explanation targets, and limitations. In addition, we discuss current evaluation practices, highlight key challenges in achieving reliable and deployable XAI-HAR, and outline directions toward trustworthy activity recognition systems that better support human understanding and decision-making.
\end{abstract}

\begin{IEEEkeywords}
Human activity recognition, explainable artificial intelligence, sensor-based activity recognition, model interpretability, deep learning, trustworthy machine learning.
\end{IEEEkeywords}

\section{Introduction}
\label{sec1}

\IEEEPARstart{H}{uman} activity recognition (HAR) has become a foundational component of intelligent systems across healthcare monitoring \cite{davidashvilly2022activity, khaliq2023decoding}, smart homes \cite{fiori2024gnn, das2023explainable}, human-robot collaboration \cite{benos2025explainable}, sports analytics \cite{hendry2020development}, and assistive technologies \cite{robinson2023deep}. Advances in wearable sensors, ambient IoT infrastructures, and multimodal sensing platforms have enabled continuous observation of human behavior through high-dimensional, multivariate time-series data. In parallel, machine learning, particularly deep learning, has driven substantial gains in recognition accuracy by modeling complex temporal dependencies, cross-sensor interactions, and multimodal correlations. These performance improvements, however, have come at the cost of transparency, raising concerns about trust, reliability, and accountability when HAR systems are deployed in real-world, often safety-critical, environments.

Explainable artificial intelligence (XAI) addresses this challenge by developing methods that make the behavior and decisions of machine learning models understandable to humans. Broadly, XAI seeks to answer why a model produces a particular prediction, which inputs or internal representations influence that decision, and how the model behaves across operating conditions or data distributions \cite{doshi2017towards, molnar2020interpretable}. Early work on interpretability focused on inherently transparent models such as linear classifiers, decision trees, and probabilistic graphical models, where explanations could be derived directly from model parameters or decision paths \cite{bastani2017interpretability, gilpin2018explaining, kim2018interpretability, ismail2020benchmarking}. While these approaches offer interpretability by design, they lack the expressive power required to model complex perceptual and temporal phenomena, including those encountered in realistic HAR settings.

The widespread adoption of deep learning fundamentally altered this landscape. Convolutional, recurrent, and hybrid neural architectures achieved state-of-the-art performance across vision, speech, and sequential modeling, but introduced highly non-linear and opaque decision processes. This shift motivated the development of post-hoc explanation techniques designed to analyze trained black-box models without modifying their internal structure. Early gradient-based saliency methods visualized input sensitivities by computing derivatives of the output with respect to the input \cite{simonyan2013deep, springenberg2015striving}, while perturbation-based approaches approximated local decision boundaries using simplified surrogate models \cite{ribeiro2016should}. These ideas were later formalized into influential frameworks such as Local Interpretable Model-agnostic Explanations (LIME) and SHapley Additive exPlanations (SHAP), which provide model-agnostic feature attribution with theoretical guarantees grounded in local fidelity and cooperative game theory \cite{ribeiro2016should, lundberg2017unified}. In parallel, model-specific techniques including Integrated Gradients (IG), Class Activation Mapping (CAM), and Grad-CAM leverage internal representations to generate spatially or temporally coherent explanations \cite{sundararajan2017axiomatic, selvaraju2020grad}. More recent XAI research has moved beyond feature attribution toward concept-based reasoning, counterfactual explanations, and human-centered explanation frameworks that explicitly consider deployment constraints, user roles, and decision-support objectives \cite{kim2018interpretability, wachter2017counterfactual, miller2019explanation, gunning2019xai, arrieta2020explainable, bhatt2020explainable, rawal2021recent, sil2025challenges, rong2023towards}.

Despite this progress, existing survey literature on explainable AI in HAR (XAI-HAR) remains fragmented. Many reviews conflate conceptual dimensions of explainability—such as local versus global explanations or intrinsic versus post-hoc interpretability—with algorithmic mechanisms, leading to overlapping or loosely defined taxonomies. Attribution-based methods dominate existing surveys, while graph-based, generative, and concept-driven approaches are less consistently represented or systematically discussed. Evaluation strategies are often addressed at a high level, with limited emphasis on explanation faithfulness, stability, and the temporal characteristics specific to HAR data. Furthermore, several surveys treat HAR explainability primarily as an extension of computer vision, without fully accounting for the distinctive challenges posed by multivariate sensor time-series and activity semantics. This review addresses these gaps through a structured and integrative synthesis of the XAI-HAR literature. We introduce a unified framework that organizes methods across complementary interpretability dimensions, alongside an algorithmic taxonomy that categorizes approaches by their dominant explanation mechanisms. By separating conceptual dimensions from algorithmic framework, the study reduces ambiguity while preserving methodological diversity. We further provide a taxonomy-consistent comparison that characterizes each method category in terms of interpretability objectives, explanation targets, and key limitations in HAR, and a technological synthesis linking conceptual and algorithmic perspectives to address temporal, multimodal, and semantic complexities. Moreover, we identify key open challenges that limit the reliability and practical deployment of XAI-HAR systems. The paper is organized as follows. Section~\ref{sec2} introduces XAI in the context of HAR and outlines common system-level design patterns. Section~\ref{sec3} presents the conceptual framework, followed by Section~\ref{sec4}, which discusses the algorithmic taxonomy. Section~\ref{sec5} provides a technological summary and comparative discussion, including open challenges. Finally, Section~\ref{sec6} concludes with future research directions.

\section{Explainable AI in Human Activity Recognition}
\label{sec2}

Explainable artificial intelligence in HAR refers to the systematic integration of interpretability mechanisms that enable humans to understand, validate, and trust activity predictions derived from sensor-based machine learning models \cite{uddin2021human, pellano2024movements, benos2025explainable}. In contrast to generic feature-importance analysis, explainability in HAR must clarify how temporal sensor dynamics, multimodal interactions, and contextual cues jointly give rise to activity decisions. This requirement is particularly critical in healthcare monitoring, assisted living, and human–robot collaboration, where explanations serve both technical validation and user-facing interpretability functions.

\begin{figure*}[!t]
\centering
\includegraphics[width=5in]{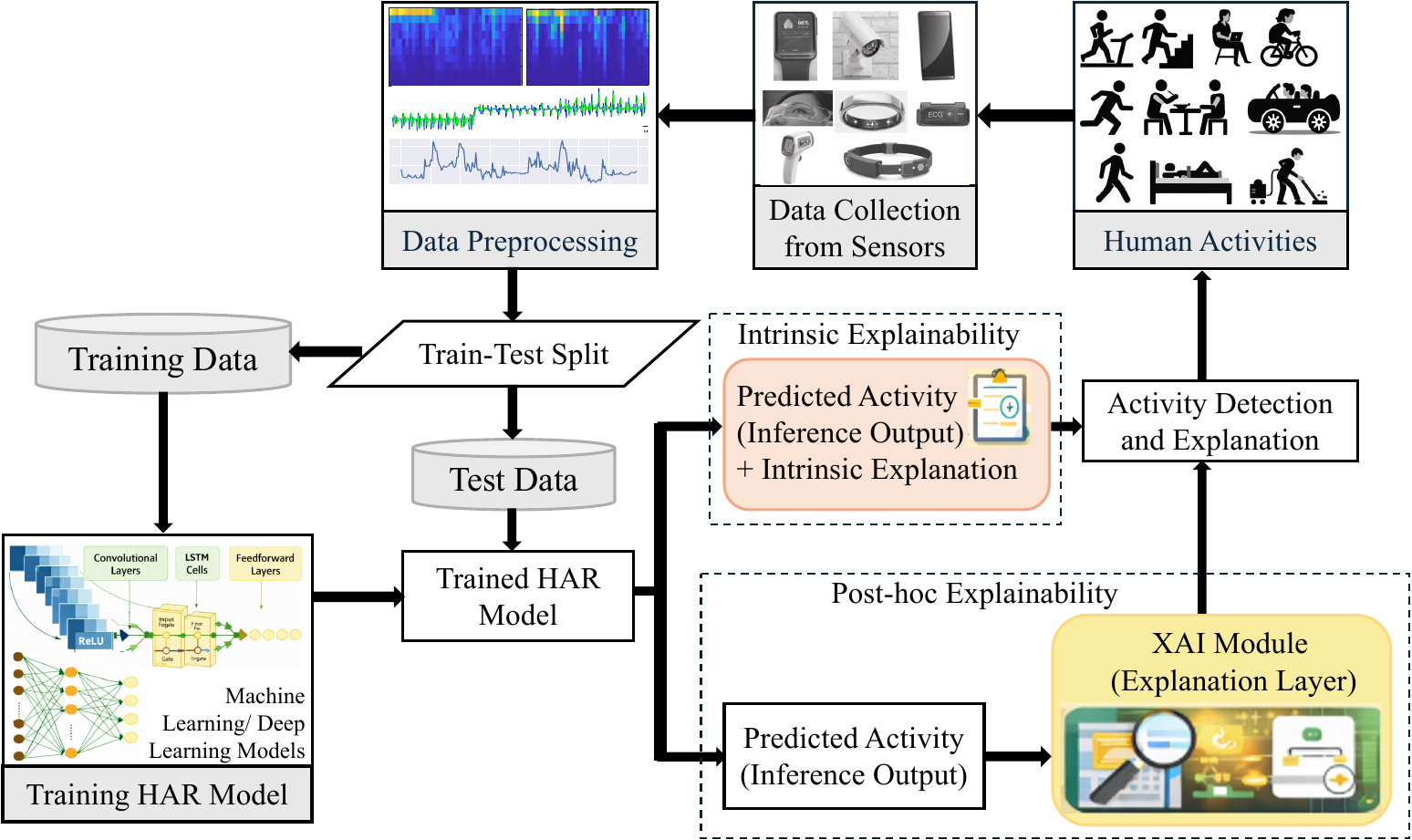}
\caption{Canonical explainable HAR pipeline illustrating the separation between training and inference and two complementary explainability paradigms. During inference, explainability is realized either intrinsically within the HAR model or post-hoc via an external XAI module, with both paths converging at a unified activity detection and explanation stage.}
\label{fig1}
\end{figure*}

A defining characteristic of XAI-HAR lies in the nature of its data. HAR relies primarily on multivariate time-series streams collected from wearable inertial measurement units, physiological sensors such as electroencephalography (EEG), vision-derived skeleton representations, and ambient smart-home sensors \cite{kanjilal2022rich, kanjilal2025human}. These signals are inherently noisy, strongly correlated, and only indirectly related to semantic activity labels \cite{arul2024revealing, liu2024enhanced}. Unlike image or text domains where raw input can be directly inspected, HAR explanations must bridge the semantic gap between low-level sensor measurements and high-level human actions \cite{pellano2024movements}. This challenge has motivated explanation strategies that explicitly account for temporal structure, modality interactions, and contextual dependencies rather than treating inputs as independent features \cite{lara2012survey, ramanujam2021human, kanjilal2022rich, jeyakumar2023x, das2023explainable, garcia2023deep, kanjilal2025human}. Fig.~\ref{fig1} illustrates a canonical explainable HAR pipeline that reflects a widely adopted design paradigm rather than a rigid standard. The pipeline explicitly separates the training and inference phases and highlights two complementary explainability strategies applied during inference \cite{sundararajan2017axiomatic, selvaraju2017grad, das2023explainable, hussain2023explainable, arul2024revealing, liu2024enhanced}. After sensor data collection and preprocessing, the dataset is split into training and test subsets, with machine learning or deep learning models trained exclusively on the training data \cite{kanjilal2022rich, arul2024revealing, liu2024enhanced, kanjilal2025human}. During inference, test data are processed by the trained HAR model, after which explainability is realized through one of two paths. In intrinsically explainable HAR models, predictions and explanations are generated jointly within the model architecture, yielding activity recognition outcomes accompanied by built-in interpretability. In contrast, post-hoc explainability applies an external XAI module to analyze the trained model’s behavior and inference output, producing explanations after prediction using attribution- and perturbation-based techniques commonly adopted in deep time-series models \cite{sundararajan2017axiomatic, selvaraju2017grad}.

\begin{figure*}[!t]
\centering
\includegraphics[width=6in]{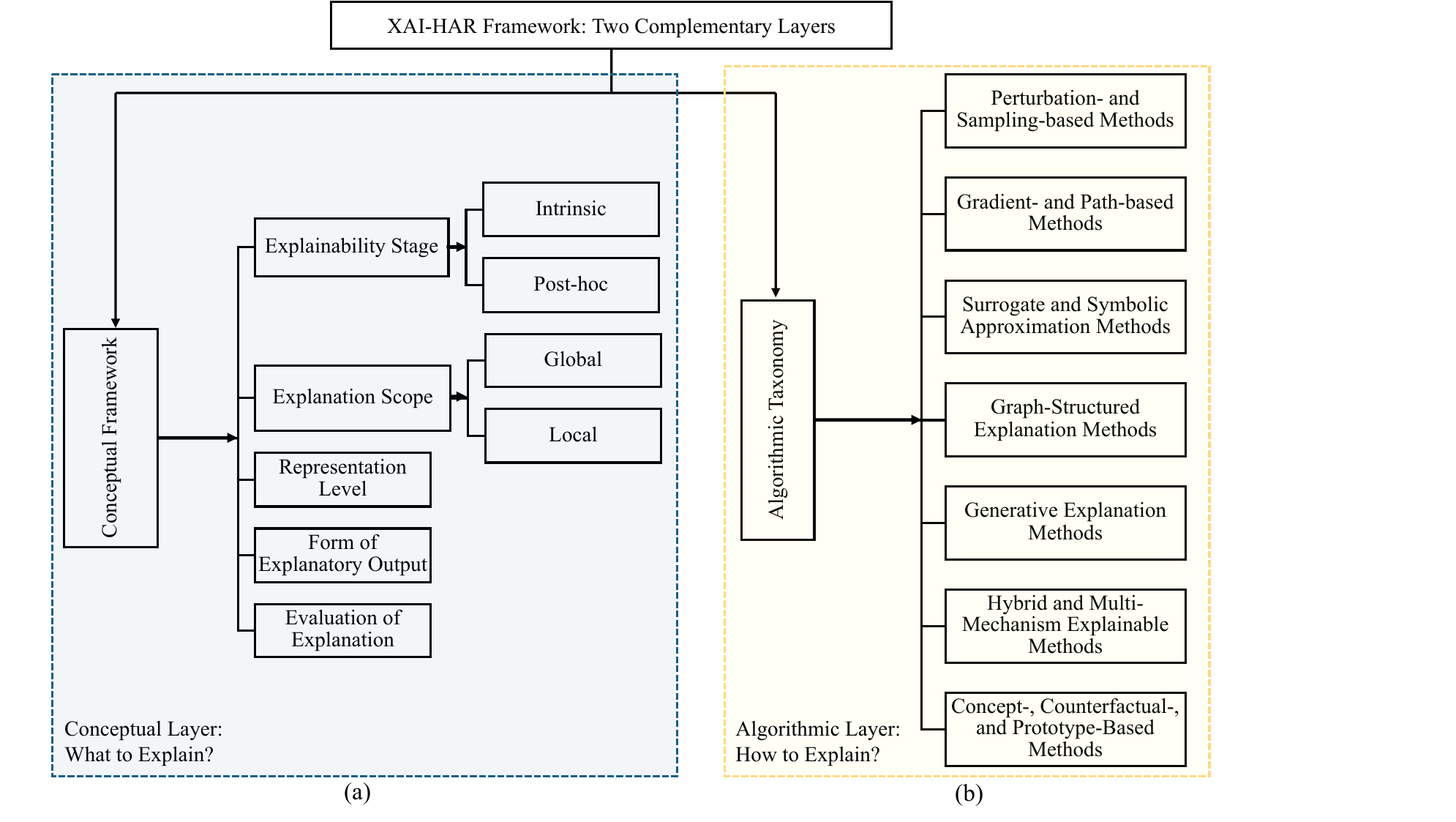}
\caption{Structural overview of XAI-HAR illustrating two complementary layers: (a) the conceptual layer, which defines what aspects of model behavior should be explained; and (b) the algorithmic layer, which categorizes how explanations are generated through different families of methods. The conceptual layer guides the selection and design of algorithmic approaches.}
\label{fig2}
\end{figure*}

Importantly, XAI-HAR systems often support multiple forms of explanation simultaneously to address diverse user requirements. Feature-level explanations assist developers in diagnosing misclassifications and detecting spurious correlations \cite{arul2024revealing}, whereas higher-level semantic or structural explanations enable clinicians, caregivers, and system operators to assess alignment between model behavior and domain knowledge \cite{jeyakumar2023x, sun2023gexse, fiori2024gnn}. In smart-home and assistive environments, explanations can be translated into visual reconstructions or natural-language descriptions to enhance transparency and trust \cite{das2023explainable, de2025sez}. To systematically organize this diverse landscape, Fig.~\ref{fig2} presents a structural overview of XAI-HAR that separates interpretability into two complementary layers. The first layer establishes a conceptual framework that defines the fundamental dimensions of explainability, while the second layer provides an algorithmic taxonomy that categorizes the computational mechanisms used to generate explanations. This distinction prevents conflating interpretability objectives with implementation strategies and clarifies methodological differences in the literature \cite{lara2012survey, ramanujam2021human, kanjilal2022rich, jeyakumar2023x, das2023explainable, garcia2023deep, kanjilal2025human}.

\section{Conceptual Framework of XAI-HAR}
\label{sec3}

XAI-HAR is inherently multidimensional, shaped by diverse design and evaluation considerations. Unlike vision or language tasks, HAR relies on multivariate time-series sensor data that are noisy, high-dimensional, and difficult to interpret directly. As a result, explainable HAR methods must not only clarify model behavior but also translate low-level sensor patterns into representations aligned with human reasoning. To structure this space, we define five complementary dimensions for characterizing the conceptual framework, which is shown in Fig.~\ref{fig2}(a).

\subsection{Explainability Stage: Intrinsic and Post-hoc Interpretability}

The explainability stage refers to when interpretability is introduced in the modeling pipeline. In HAR, methods broadly fall into intrinsic and post-hoc approaches. intrinsic explainability embeds interpretability directly into the model architecture. For instance, explainable complex human activity recognition (X-CHAR) model introduces concept-based reasoning for complex activities \cite{jeyakumar2023x}, GNN-XAR derives explanations from graph-structured sensor representations \cite{fiori2024gnn}, and GeXSe (Generative Explanatory Sensor System) integrates generative reconstruction into the recognition pipeline \cite{sun2023gexse}. While intrinsic methods yield more faithful explanations, they often require carefully designed representations and are less easily transferable across datasets. In contrast, post-hoc explainability dominates current HAR research due to its flexibility and compatibility with high-performing but opaque models such as CNNs and LSTMs. These methods analyze trained models retrospectively to estimate feature relevance, generate saliency or activation maps, extract rules, or produce human-readable explanations without modifying the underlying classifier. In \cite{aquino2023explaining}, the authors analyzed latent embeddings of CNN-based HAR models to understand activity separability, while in another study \cite{hussain2023explainable}, the authors applied LIME to EEG-based HAR to identify influential channels and frequency bands. 

\subsection{Explanation Scope: Global and Local}

Explanation scope distinguishes whether interpretability targets overall model behavior or individual predictions. Global explanations are valuable for model validation, sensor selection, and bias analysis. In \cite{liu2024enhanced, benos2025explainable}, global feature importance analysis was used to identify dominant sensors and inform system-level design decisions. More recent work has emphasized quantitative global evaluation, with faithfulness and stability assessed in skeleton-based HAR settings \cite{pellano2024movements}. Local explanations, on the other hand, focus on why a specific activity instance was predicted. In HAR, local reasoning is critical due to strong temporal dependencies and subject variability. Attribution-based explanations, such as LIME \cite{hussain2023explainable} and Grad-CAM \cite{mekruksavanich2025efficient}, highlight influential temporal segments or sensor channels, while generative methods such as SEZ-HARN (Self-Explainable Zero-shot Human Activity Recognition Network) produce instance-specific motion visualizations \cite{de2025sez}. 

\subsection{Representation Level of Explanations}

Explanations in HAR vary in abstraction, ranging from feature-level to semantic and generative representations. Feature-level explanations remain most common, attributing importance to sensor channels, time windows, or frequency components \cite{hussain2023explainable, liu2024enhanced, mekruksavanich2025efficient}. However, such explanations often require domain expertise. Higher-level representations aim to bridge this gap. Concept-based explanations, as in \cite{jeyakumar2023x}, align model reasoning with human-interpretable sub-activities, while structural explanations, such as those in \cite{fiori2024gnn}, capture relationships among sensors or entities. At the highest abstraction level, generative explanations in \cite{sun2023gexse, de2025sez} translate internal representations into visual or motion-based narratives, shifting explainability from diagnostic analysis toward human-centric communication.

\subsection{Forms of Explanatory Outputs}

Explainability methods in HAR also differ in the form of their outputs. Attribution maps (saliency, activation heatmaps) remain the most prevalent, particularly for deep CNN-based models. Rule-based and symbolic explanations, including concept chains or natural-language narratives, provide structured and human-readable justifications, as demonstrated in X-CHAR \cite{jeyakumar2023x} and smart-home HAR studies \cite{das2023explainable}. More expressive outputs include visual and multimodal narratives, such as generated activity scenes or graph visualizations, which facilitate trust and understanding among end users \cite{sun2023gexse, fiori2024gnn}.

\subsection{Evaluation of Explainability}

Evaluating explainability is a persistent challenge in XAI-HAR. While qualitative inspection remains common, recent work emphasizes quantitative criteria such as faithfulness, stability, and consistency. Benchmark metrics for skeleton-based HAR were introduced in \cite{pellano2024movements}, the consistency of SHAP explanations across activities was analyzed in \cite{benos2025explainable}, and similarity measures for generated explanations were proposed in \cite{de2025sez}. These efforts highlight the need for standardized evaluation protocols tailored to HAR’s temporal and multimodal nature. 

\section{Algorithmic Taxonomy of XAI-HAR}
\label{sec4}

The conceptual dimensions introduced in Section~\ref{sec3} directly motivate the algorithmic taxonomy presented in this section. In the context of HAR, explainability is not a single objective but a collection of complementary questions concerning temporal evidence, sensor and modality contributions, and semantic activity structure. Accordingly, the methods reviewed in Section~\ref{sec4} are not only categorized by their underlying mechanisms but can also be interpreted by the specific HAR-related explanation needs they address. As shown in Fig.~\ref{fig2}(b), perturbation and gradient-based methods primarily offer local, instance-level insights into temporal or feature sensitivity, whereas surrogate, concept-based, and prototype-driven approaches emphasize class-level or semantic explanations of activity decisions. Graph-structured and hybrid methods further attempt to capture relational and multi-level aspects of human motion, reflecting the inherently structured nature of HAR data. This perspective enables a clearer understanding of both the strengths and the limitations of existing XAI approaches when applied to HAR.

\subsection{Perturbation- and Sampling-Based Methods}

Perturbation- and sampling-based methods explain model behavior by systematically altering input features and observing the resulting changes in model predictions. These approaches operate independently of model internals and are therefore broadly applicable to black-box HAR systems. 

\subsubsection{Local Interpretable Model-Agnostic Explanations}

LIME is a perturbation-based, post-hoc explainability technique designed to provide local, instance-level explanations for predictions made by black-box classifiers \cite{ribeiro2016should}. Let $f(x)$ denote a trained black-box model and $x \in \mathbb{R}^d$ an input instance whose prediction is to be explained. LIME constructs a perturbed dataset $\{z_i\}$ around $x$ by randomly modifying components of the input, such as masking or altering feature segments, while preserving the overall structure of the instance. Each perturbed sample is weighted according to its proximity to the original instance through a locality kernel $\pi_x(z)$, which ensures that samples closer to $x$ have greater influence on the explanation. An interpretable surrogate model $g \in \mathcal{G}$, typically a sparse linear model, is then learned by solving the optimization problem:

\begin{equation}
    g^* = \arg\min_{g \in \mathcal{G}} \; L(f, g, \pi_x) + \Omega(g),
\label{eq1}
\end{equation}

\noindent where $L(f, g, \pi_x)$ measures the fidelity of the surrogate model to the predictions of the black-box classifier in the local neighborhood of $x$, commonly implemented via a locality-weighted least-squares loss. In \eqref{eq1}, $\Omega(g)$ is a complexity regularizer that enforces interpretability by constraining the number of non-zero coefficients in $g$. The coefficients of the optimized surrogate model $g^*$ provide feature-attribution scores that indicate which input components most strongly influenced the prediction $f(x)$. By construction, these explanations are locally faithful but are not intended to generalize beyond the neighborhood of the explained instance \cite{liu2024enhanced}.

\begin{figure*}[!t]
    \centering
    \includegraphics[width=6in]{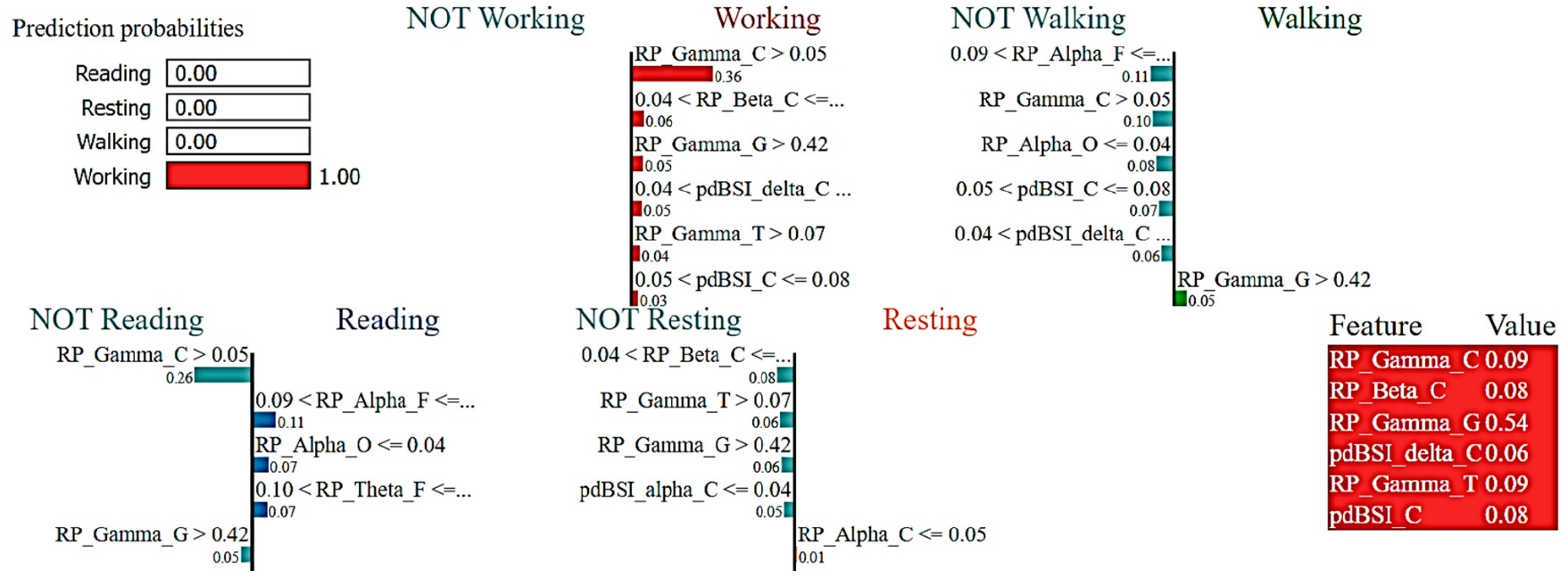}
    \caption{Visualization of the local contribution of EEG features using the LIME model for classifying a single test instance (predicted class = Working activity) with the Random Forest model. The red-marked cells represent the features that contributed most to the classification of the Working activity \cite{hussain2023explainable}.}
\label{fig3}
\end{figure*}

In HAR, LIME is adapted to multivariate time-series by defining interpretable features as temporal segments, sensor windows, or frequency components. In \cite{uddin2021human}, LIME was applied to explain predictions of a wearable-sensor–based HAR system. After training an LSTM-based model on multimodal inertial and physiological features, LIME generated local explanations by perturbing temporal sensor-window representations. The resulting surrogate models highlighted the most influential sensor-derived features for each activity decision, such as walking or running, providing instance-level interpretability. In another study, Hussain et al. extended LIME to EEG-based HAR by applying perturbations to channel-wise and frequency-domain EEG features, enabling the identification of neurophysiologically relevant signal components contributing to activity recognition \cite{hussain2023explainable}. While LIME explanations highlighted meaningful EEG channels and spectral bands, the authors reported that explanation stability is affected by strong inter-channel correlations and high-dimensional feature spaces, underscoring challenges inherent to perturbation-based attribution in biosignal analysis. As illustrated in Fig.~\ref{fig3}, LIME highlights the dominant EEG spectral features contributing to a correctly classified “working” activity instance. The visualization shows that central and global gamma-band features receive the highest attribution weights (up to 36\%), while central beta-band features contribute negatively, reflecting the model’s reliance on cognitively demanding motor-related neural patterns. Although the explanation aligns well with known neurophysiological interpretations, Fig.~\ref{fig3} also demonstrates that attribution weights are highly instance-specific, reinforcing the authors’ observation that LIME explanations can vary under correlated EEG channels and high-dimensional feature spaces. 

Bijalwan et al. \cite{bijalwan2024interpretable} further examined this locality-stability trade-off in an inertial measurement unit (IMU)-based setting by pairing LIME with a temporal convolutional network (TCN) pipeline and a multimodal fusion model that combines TCN, CNN, and LSTM branches for seven discrete activities. Using an in-house dataset, they reported high performance (mean accuracy of approximately 98.7\%) and subsequently applied LIME to fit local surrogate models around individual test instances, ranking sensor-derived features according to their contribution to each activity decision. Their explanations emphasize that orientation-related features (Angle X/Y/Z) and time are consistently influential, while acceleration components contribute less. Importantly, the study demonstrates that LIME can surface discriminative feature ranges for specific activities (e.g., particular angle intervals supporting \emph{downstairs}/\emph{normal walking} versus angle Y intervals supporting \emph{jogging}/\emph{upstairs}). However, the authors also note that correlated IMU features and noise-handling steps can cause explanation patterns to vary across instances, echoing broader concerns regarding the stability of perturbation-based explanations on correlated wearable streams \cite{bijalwan2024interpretable}. Similarly, in a recent study \cite{liu2024enhanced}, the authors integrated LIME within a wearable HAR framework. In their CNN--LSTM-based system, which was trained on the UCI HAR dataset, LIME was employed specifically for local explanation of individual predictions. Their findings reinforce the view that LIME is particularly effective for sample-level debugging and post-hoc inspection of misclassifications.

A more systematic evaluation of LIME in ambient sensor-based HAR was conducted in \cite{das2023explainable}, where the authors proposed a human-centered XAI-HAR framework for smart-home environments and applied LIME to explain LSTM-based activity recognition models operating on multivariate ambient sensor data. In this pipeline, raw smart-home sensor streams are transformed into fixed-interval features and classified using an LSTM-based HAR model, followed by a post-hoc explanation module. Using LIME, the authors perturbed binary and temporal sensor features to identify influential sensor activations and generate natural-language explanations. To address limitations in handling long-duration and temporally correlated activities, they proposed LIME+, which aggregates feature importance over time to capture sustained sensor relevance. As a result, LIME+ produces more coherent and interpretable explanations by emphasizing persistent sensor contributions, while retaining local faithfulness to the HAR model.

\subsubsection{Kernel Shapley Additive Explanations}

Building on perturbation-based attribution methods, SHAP provides a principled framework for feature attribution based on Shapley values. It assigns each input feature an importance value for a specific model prediction. Formally, SHAP represents explanations using an additive feature attribution model of the form \cite{lundberg2017unified}:

\begin{equation}
    g(z') = \phi_0 + \sum_{i=1}^{M} \phi_i z'_i,
\label{eq2}
\end{equation}

\noindent where $z' \in \{0,1\}^M$ denotes a simplified binary representation of the input, $\phi_i$ represents the contribution of feature $i$, and $\phi_0$ corresponds to the expected model output in the absence of all features. The attribution values $\phi_i$ are computed as Shapley values, defined as the weighted average marginal contribution of a feature over all possible feature subsets 

\begin{equation}
    \phi_i = \sum_{S \subseteq F \setminus \{i\}} 
    \frac{|S|!(|F| - |S| - 1)!}{|F|!}
    \left[ f_{S \cup \{i\}}(x_{S \cup \{i\}}) - f_S(x_S) \right].
\label{eq3}
\end{equation}

\noindent In \eqref{eq3}, $F$ denotes the set of all input features with $|F| = M$, and $S \subseteq F \setminus \{i\}$ represents a subset of features excluding feature $i$. The term $|S|$ denotes the cardinality of $S$. The functions $f_{S}(x_{S})$ and $f_{S \cup \{i\}}(x_{S \cup \{i\}})$ denote the model outputs when only the features in $S$ or $S \cup \{i\}$ are present, respectively. The weighting factor 
$\frac{|S|!(|F| - |S| - 1)!}{|F|!}$ corresponds to the Shapley value coefficient, which ensures a fair averaging of the marginal contribution of feature $i$ over all possible feature subsets \cite{lundberg2017unified}. SHAP provides the unique additive feature attribution method that satisfies the desirable properties of local accuracy, missingness, and consistency, thereby unifying several existing explanation techniques under a common theoretical framework. In practice, SHAP values are approximated using efficient algorithms such as Kernel SHAP for model-agnostic settings and TreeSHAP or Deep SHAP for specific model classes.

\begin{figure}[!t]
    \centering
    \includegraphics[width=3.5in]{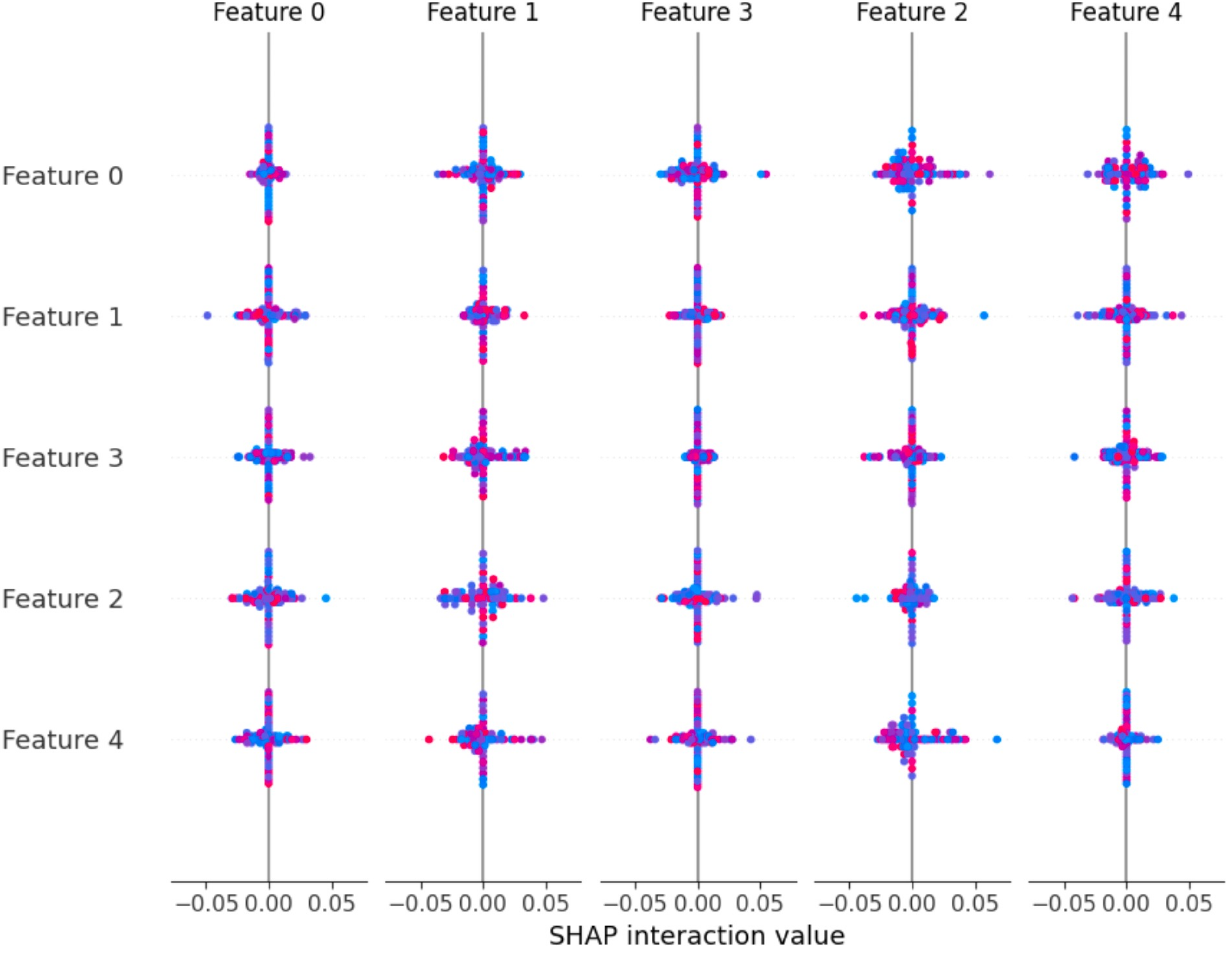}
    \caption{SHAP interaction plot showing the most influential features and their pairwise interactions affecting CNN–LSTM–based activity predictions \cite{liu2024enhanced}.}
\label{fig4}
\end{figure}

In HAR, SHAP has been widely adopted to provide both global and local explanations of complex sensor-based models. In \cite{liu2024enhanced}, the authors incorporated SHAP into an explainable wearable HAR framework based on a CNN–LSTM architecture trained on the UCI HAR dataset to analyze how tri-axial accelerometer features influence activity predictions. After training the CNN–LSTM model, SHAP was applied using a kernel-based explainer to compute shapley values that quantify each feature’s contribution to the model’s output while preserving consistency and fairness guarantees. Fig.~\ref{fig4} visualizes both individual feature importance and pairwise feature interactions, revealing that a small subset of features (features 0-4) dominates the model’s predictions, with strong interaction effects among specific feature pairs \cite{liu2024enhanced}. The distribution of SHAP interaction values illustrates how certain features consistently increase or decrease prediction confidence across samples, while others exhibit weaker or more variable influence. Through this analysis, the authors demonstrate that SHAP enables interpretable insights into both dominant features and their interactions in time-series sensor data, supporting global understanding of model behavior as well as instance-level diagnosis of activity recognition decisions. In another study \cite{arul2024revealing}, SHAP was integrated with an XGBoost classifier trained on the UCI HAR dataset to analyze how time- and frequency-domain features derived from smartphone accelerometer and gyroscope signals influence activity predictions. Using SHAP values, the authors quantified both instance-level feature contributions and overall feature importance across the dataset, enabling a clear understanding of how individual sensor features positively or negatively affect specific activity classes. Bijalwan et al. similarly employed SHAP to analyze feature importance in TCN-based HAR models, highlighting its effectiveness in capturing long-range temporal dependencies through global attribution scores \cite{bijalwan2024interpretable}.

A more application-oriented use of SHAP was presented by Benos et al., who employed TreeSHAP to explain an XGBoost-based HAR model developed for human–robot collaboration in agricultural environments \cite{benos2025explainable}. SHAP quantified the contribution of multi-location IMU channels, highlighting the dominant role of torso-mounted magnetometer features in capturing core movements. The analysis also guided sensor placement and feature relevance, demonstrating how explainability can enhance efficiency, safety, and trust in real-world HAR systems. In a clinical context, SHAP was applied to CNN–LSTM models for cognitive health assessment, using global feature attributions to support population-level interpretation of movement and physiological patterns relevant to cognitive decline \cite{javed2023toward}. In \cite{das2023explainable}, SHAP was evaluated for ambient sensor–based HAR in smart-home settings. In their study, SHAP was applied alongside LIME and Anchors to LSTM models trained on long, event-driven sequences. It provided both local and global explanations, capturing feature importance across extended temporal contexts. SHAP produced more temporally coherent and domain-consistent explanations than LIME, highlighting its suitability for long-duration smart-home HAR and enhancing user trust. More recently, Tempel et al. proposed ShapGCN for skeleton-based HAR, applying SHAP to graph convolutional networks and validating explanations through targeted joint perturbations. Their results show strong alignment between SHAP-attributed joint importance and model sensitivity, confirming the fidelity of SHAP explanations for graph-based HAR models \cite{tempel2025explaining}.

\subsubsection{Anchors}

Anchors provide high-precision, rule-based explanations by identifying feature predicates that reliably ``anchor'' a prediction under perturbation-based sampling \cite{ribeiro2018anchors}. Although explanations are expressed symbolically, the underlying mechanism relies on probabilistic perturbation and sampling strategies, justifying its placement within this category. Anchors explain a prediction $f(x)$ by identifying a rule $A$, defined as a set of feature predicates over an interpretable representation, such that the model’s prediction remains unchanged with high probability whenever the rule holds \cite{ribeiro2018anchors}. Let $D(z)$ denote a perturbation distribution over the input space and $D(z \mid A)$ the conditional distribution of samples satisfying rule $A$. Rule $A$ is considered an anchor for instance $x$ if it satisfies

\begin{equation}
    \mathbb{E}_{z \sim D(z \mid A)} 
    \left[ \mathbf{1}_{(f(z) = f(x))} \right] \ge \tau,
\label{eq4}
\end{equation}

\noindent where $\tau$ is a user-defined precision threshold and $\mathbf{1}(\cdot)$ is the indicator function. Since the true precision cannot be computed exactly, Anchors impose a probabilistic guarantee:

\begin{equation}
    \mathbb{P}\left( \mathrm{prec}(A) \ge \tau \right) \ge 1 - \delta,
\label{eq5}
\end{equation}

\noindent where $\delta$ controls the confidence level of the precision estimate. Among all rules satisfying this constraint, Anchors select the one with maximum coverage, defined as

\begin{equation}
    \mathrm{cov}(A) = \mathbb{E}_{z \sim D(z)} \left[ A(z) \right].
\label{eq6}
\end{equation}

\noindent This leads to the following optimization problem:

\begin{equation}
    \max_{A} \ \mathrm{cov}(A)
    \quad \text{s.t.} \quad
    \mathbb{P}\left( \mathrm{prec}(A) \ge \tau \right) \ge 1 - \delta.
\label{eq7}
\end{equation}

\noindent The search for such rules is performed efficiently using a perturbation-based strategy formulated as a multi-armed bandit problem, yielding short, human-interpretable if--then rules that prioritize precision and provide explicitly defined coverage. In \cite{das2023explainable}, the authors applied Anchors to explain LSTM-based activity recognition models trained on CASAS smart-home datasets, using the method to derive high-precision, rule-based explanations in the form of minimal if--then conditions over sensor activations. Their study showed that Anchors generate intuitive, temporally grounded rules composed of specific combinations of motion, door, and appliance sensors that align well with domain knowledge of activities of daily living and the physical layout of the home.

\subsubsection{Permutation Feature Importance} 

Permutation-based attribution quantifies global feature importance by measuring prediction degradation under systematic feature shuffling. As this method relies on input perturbation rather than model internals, it naturally aligns with the perturbation-based explanation paradigm. Permutation-based attribution methods have been explored in HAR to provide global interpretability of sensor contributions. Tokas et al. employed permutation feature importance to quantify the relevance of sEMG muscle channels and inertial sensor inputs in a lightweight multihead CNN–LSTM (X-LiteHAR) model \cite{tokas2025lightweight}. By measuring performance degradation under systematic feature shuffling, their analysis identified dominant sensing modalities and muscle activation patterns, supporting both model simplification and clinically meaningful interpretation. 

\subsection{Gradient- and Path-Based Methods}

Gradient- and path-based methods explain HAR model predictions by attributing relevance directly through model derivatives, leveraging gradients or integrated paths between reference inputs and observed signals. These approaches exploit internal model representations to produce locally faithful, temporally coherent explanations without relying on explicit input perturbations.

\subsubsection{Saliency Maps}

Saliency maps estimate feature relevance by computing the gradient of the model output with respect to the input, thereby identifying which temporal segments or sensor channels most strongly influence a prediction. Although computationally efficient for differentiable models, vanilla saliency methods are sensitive to noise and gradient saturation, often producing unstable attribution patterns. In HAR, saliency-based explanations have been applied across diverse sensing modalities. Aquino et al. demonstrated that gradient attribution can extend beyond raw accelerometer inputs to latent CNN embeddings, linking separability of activities to internal feature representations \cite{aquino2023explaining}. Similar visualization strategies have been explored in audio-based HAR, where gradient-based saliency highlights discriminative acoustic segments corresponding to activity-specific sound patterns \cite{kim2025granular}, and in radar-based HAR, where salient micro-Doppler signatures and temporally localized motion dynamics are identified within spectrogram representations \cite{waghumbare2025diat}. Despite their flexibility, the instability of local gradients motivated more principled aggregation methods, most notably IG.

\subsubsection{Integrated Gradients}

IG explains individual predictions of differentiable models by quantifying how each input feature contributes to the output along a continuous path from a reference baseline to the actual input. IG was designed to address gradient saturation while satisfying axioms such as sensitivity and implementation invariance \cite{sundararajan2017axiomatic}. Let $f(x)$ denote the pre-softmax score of a target activity class for an input $x$, and let $x'$ be a baseline input (e.g., zero signal or mean activity). The attribution assigned to the $i$-th input feature is defined as

\begin{equation}
    \mathrm{IG}_i(x) = (x_i - x'_i)
    \int_{0}^{1}
    \frac{\partial f\!\left(x' + \alpha (x - x')\right)}
    {\partial x_i}
    \, d\alpha,
\label{eq8}
\end{equation}

\noindent where $\alpha \in [0,1]$ parameterizes a straight-line interpolation from the baseline $x'$ to the input $x$. By accumulating gradients along this path, IG captures the total contribution of each feature to the model’s prediction rather than relying on local gradient values at a single point. Unlike LIME and SHAP, which rely on input perturbations and surrogate modeling, IG leverages the model’s internal gradients, yielding smooth, temporally consistent attributions that align naturally with sequential sensor data \cite{ismail2020benchmarking}.

Recent HAR studies increasingly adopt IG as a complementary explanation tool. The authors in \cite{lamaakal2025tiny} incorporated IG within the XTinyHAR framework to analyze lightweight inertial Transformer models trained via multimodal knowledge distillation. Their heatmaps highlight motion-intensive temporal segments and dominant inertial channels, such as accelerometer and gyroscope responses, indicating that the student model relies on salient biomechanical motion cues for activity recognition. Importantly, IG is used alongside attention visualization and attention rollout to examine how knowledge is transferred from a multimodal teacher to a unimodal inertial student, providing complementary evidence that the distilled model preserves meaningful input–output relationships in the learned representations.

\subsubsection{Gradient-weighted Class Activation Mapping}

Grad-CAM adapts gradient-based attribution to convolutional architectures by producing class-discriminative activation maps that highlight which latent regions of convolutional feature representations most strongly influence a model’s prediction. By leveraging gradients flowing into convolutional layers, Grad-CAM exploits the internal structure of CNNs to generate explanations that are locally faithful and temporally or spatially coherent, making it particularly suitable for deep HAR models based on convolutional feature extraction. Let $f(x)$ denote the pre-softmax score of a target activity class for an input $x$, and let $A^k$ represent the $k$-th feature map of a selected convolutional layer. Grad-CAM computes the importance weight $\alpha_k^c$ of each feature map by globally averaging the gradients of the class score with respect to that feature map \cite{selvaraju2017grad}:

\begin{equation}
    \alpha_k^c =
    \frac{1}{Z}
    \sum_i \sum_j
    \frac{\partial f^c}{\partial A_{ij}^k},
\label{eq9}
\end{equation}

\noindent where $i,j$ index temporal or spatial locations and $Z$ is a normalization constant. The class-specific attribution map is then obtained as

\begin{equation}
    L_{\mathrm{Grad\text{-}CAM}}^c
    =
    \mathrm{ReLU}
    \left(
    \sum_k \alpha_k^c A^k
    \right),
\label{eq10}
\end{equation}

\noindent with the ReLU operator ensuring that only features positively contributing to the target class are visualized. This formulation generalizes CAM, which requires architectural constraints such as global average pooling and linear classification layers \cite{selvaraju2017grad}. Grad-CAM removes these constraints and can be applied to a wide range of CNN-based and hybrid CNN--RNN models commonly used in HAR.

\begin{figure*}[!t]
    \centering
    \includegraphics[width=6in]{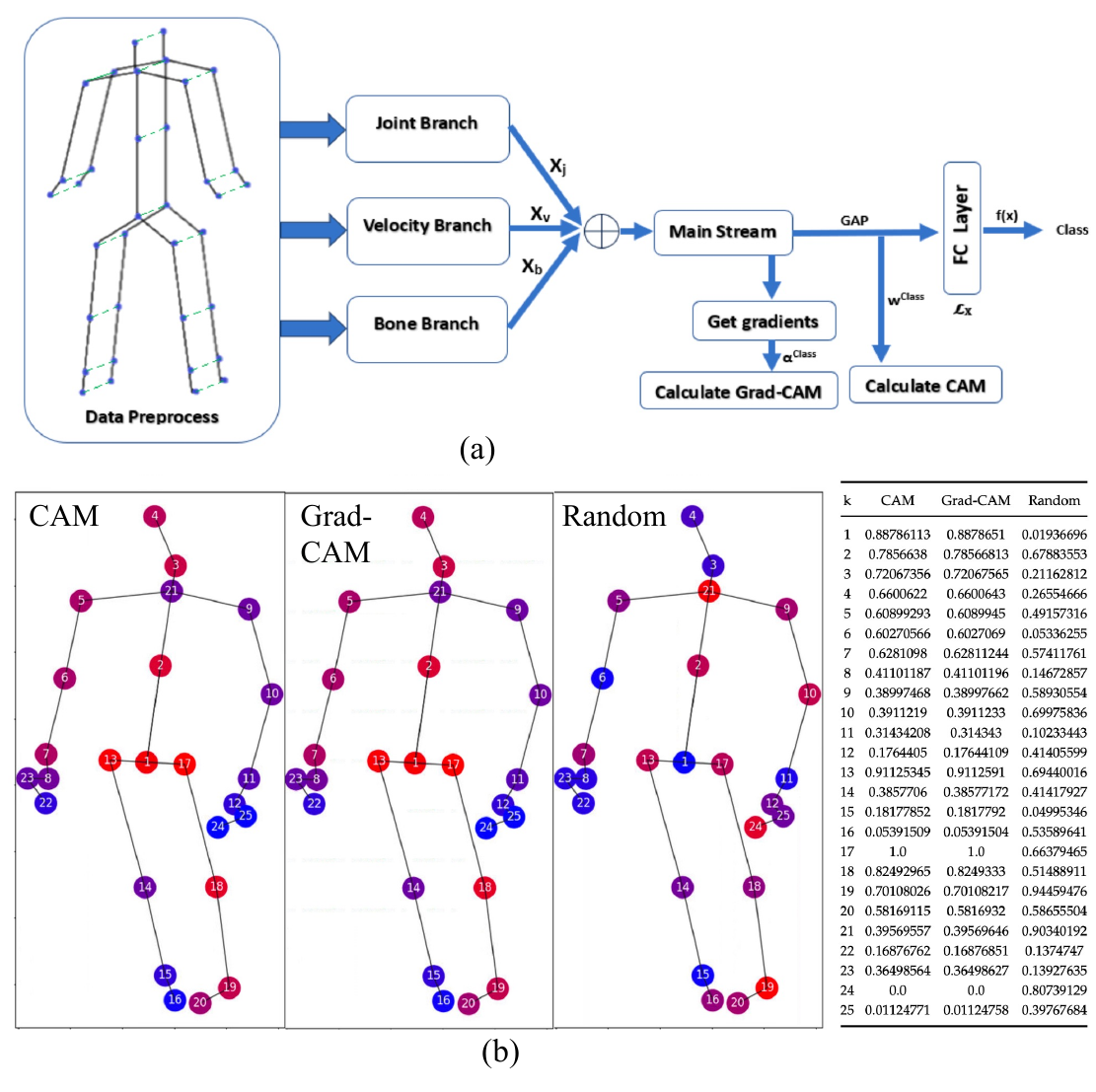}
    \caption{(a) Overview of the EfficientGCN pipeline illustrating the variables used to compute faithfulness and stability, with perturbations applied during the data preprocessing stage. (b) (Left) CAM, Grad-CAM, and random baseline attributions for a representative ‘standing up’ activity instance (class 8), averaged across all frames and normalized. The color gradient represents attribution intensity, ranging from blue (0) to red (1). (Right) Corresponding numerical attribution scores, where k denotes the body point index \cite{pellano2024movements}.}
\label{fig5}
\end{figure*}

In HAR, Grad-CAM has been widely adopted to explain deep models operating on multivariate sensor time series, where convolutional feature maps capture latent temporal patterns and sensor-channel abstractions rather than spatial image regions. In \cite{huang2024explainable}, Grad-CAM was incorporated into an explainable deep learning framework, using CNN-based architectures to visualize salient temporal segments and sensor channels that contribute to activity predictions. The visualizations provide qualitative insights into how learned representations align with intuitive motion dynamics, supporting interpretability of model decisions in time-series sensor data. In \cite{mekruksavanich2025efficient}, the authors integrated Grad-CAM into a deep residual architecture enhanced with squeeze-and-excitation mechanisms, demonstrating that Grad-CAM heatmaps can more clearly highlight discriminative temporal regions and dominant inertial channels when combined with channel-attention modules, thereby improving explanation clarity and interpretability. In \cite{pellano2024movements}, Pellano et al. conducted a comparative analysis of CAM and Grad-CAM for skeleton-based HAR, evaluating explanation quality using faithfulness and stability metrics under controlled skeletal perturbations. Fig.~\ref{fig5}(a) illustrates the experimental pipeline adopted in the study, showing how CAM and Grad-CAM explanations are extracted from an EfficientGCN-based model and assessed through systematic joint perturbations. Fig.~\ref{fig5}(b) provides qualitative joint-level visualizations for representative activity instances, demonstrating that CAM and Grad-CAM produce nearly identical relevance patterns across skeletal joints. These results indicate that the explanatory behavior of Grad-CAM is strongly influenced by the selected intermediate representation and offers limited differentiation from CAM in skeleton-based HAR. Importantly, the study shows that explanation reliability is better captured by stability under perturbations than by prediction-based faithfulness alone, highlighting limitations of relying solely on accuracy-driven metrics to assess explanation quality.

\subsection{Surrogate and Symbolic Approximation Methods}

While attribution-based methods explain model predictions by assigning relevance scores to input features, surrogate and rule-extraction methods aim to approximate the decision behavior of a complex HAR model using inherently interpretable representations, such as logical rules, decision trees, or symbolic abstractions. These methods are particularly valuable in HAR scenarios that demand high-level reasoning, semantic interpretability, and long-term behavioral understanding, such as smart-home monitoring, healthcare, and assistive environments. Unlike feature-level attribution, surrogate-based explanations emphasize decision logic and conditional structure, often at the cost of fine-grained fidelity.

\subsubsection{Decision Tree and Rule-Based Surrogates} 

Decision tree surrogates approximate complex black-box models by learning an interpretable tree that mimics their input–output behavior using queries to the trained model \cite{craven1995extracting, bastani2017interpretability}. The resulting trees provide global, hierarchical explanations through feature splits and decision paths, enabling users to trace how input features lead to specific predictions, though fidelity may degrade for highly complex models. Closely related rule-based surrogates represent model behavior as ordered if–then rules, where predictions are explained by the first satisfied rule and a default outcome otherwise \cite{angelino2018learning}. By emphasizing sparsity, simplicity, and human readability, rule-based models are particularly suitable for HAR scenarios involving clinicians, caregivers, or end users who require transparent and easily interpretable decision logic. In the HAR literature, Bettini et al. demonstrated that fully interpretable models—such as symbolic classifiers and probabilistic state-based models can act as transparent activity recognizers themselves, eliminating the need for post-hoc approximation \cite{bettini2021explainable}. Their work highlights a fundamental trade-off in HAR: while deep neural models often achieve superior recognition accuracy, decision tree– and rule-based models provide unmatched clarity, auditability, and ease of validation, which are critical in safety- and trust-sensitive applications.

\subsubsection{Logic Induction and Symbolic Reasoning}

Logic-based and rule-induction explainability justifies activity predictions through learned symbolic or temporal rules rather than feature attributions or surrogate approximations. These approaches aim to uncover high-level, human-interpretable reasoning patterns that capture causal, temporal, or structural relationships among motion primitives and sensor events. Unlike rule-based surrogates that approximate black-box classifiers post-hoc, logic-induction methods often treat symbolic rules as the primary explanatory artifact, enabling faithful and semantically grounded explanations of human activities. Another distinct paradigm is logic-based and rule-induction explainability, which justifies activity predictions through learned symbolic or temporal rules rather than feature attributions or surrogate models. The authors in \cite{cao2023discovering} introduced a framework for discovering intrinsic spatial–temporal logic rules that explain human actions. Their method automatically induces interpretable logical rules capturing causal and temporal relationships among motion primitives and demonstrates experimentally that these rules correspond to meaningful behavioral patterns. 

Logic-based explainability has also appeared in smart-home HAR through formal reasoning frameworks. In \cite{l2021probabilistic}, the authors employed probabilistic model checking to recognize activities while explicitly modeling temporal dependencies among sensor events. Although the primary goal is activity recognition, the resulting probabilistic temporal formulas provide interpretable explanations that describe why an activity is recognized over a given sensor trace. These approaches emphasize faithful, temporally grounded explanations, albeit at the cost of increased modeling complexity and reliance on structured representations.

\subsection{Graph-Structured Explanation Methods}

Graph-based explainability methods model human activities and sensor interactions as graphs, enabling explanations over nodes, edges, and their relational structure rather than flat feature vectors \cite{ying2019gnnexplainer, Pope_2019_CVPR}. Unlike feature-level attribution, which scores inputs independently, these approaches capture topology, connectivity, and dependencies in structured HAR data, such as spatial sensor relationships, temporal event ordering, and interactions among activity components. This makes them well suited for scenarios where relational structure carries semantic meaning. Explanations are typically obtained by identifying influential subgraphs, nodes, or edges through perturbation analysis of graph components, preserving relational context and revealing how interactions contribute to activity recognition decisions \cite{ying2019gnnexplainer}.

\begin{figure}[!t]
    \centering
    \includegraphics[width= 3.5 in]{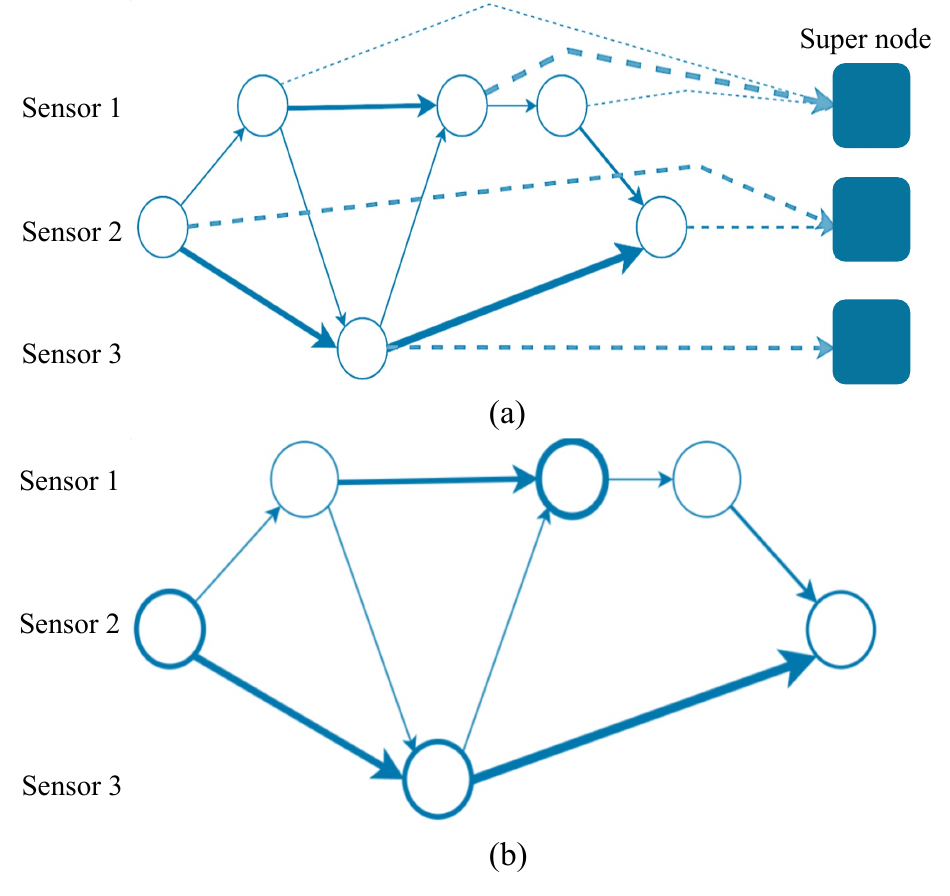}
    \caption{(a) Output of the original GNNExplainer, limited to arc importance, where the thickness of each arrow represents the importance value associated with the arc. (b) Output of the adapted version of GNNExplainer on the same input example shown in (a) \cite{fiori2024gnn}.}
\label{fig6}
\end{figure}

A HAR-specific realization of graph-based explainability was introduced through the GNN-XAR framework, which models smart-home activity data as graphs derived from ambient sensor events \cite{fiori2024gnn}. Continuous sensor streams were segmented into fixed length overlapping windows and transformed into temporal graphs encoding spatial and temporal relationships, which were processed by a graph neural network for activity classification. For explainability, GNN-XAR built on GNNExplainer \cite{ying2019gnnexplainer} to identify the subset of nodes and arcs that most strongly influenced a predicted activity by optimizing for a subgraph that maximized the mutual information between the original model prediction and the prediction obtained using only that subgraph. Fig.~\ref{fig6}(a) illustrates the arc-level importance scores produced by the original GNNExplainer, where arc thickness reflects their contribution to the prediction, including arcs connecting nodes to super-nodes used for classification. Due to the categorical nature of sensor identifiers, GNN-XAR adapts this process by extracting only arc-level importance and inferring node importance from the arcs linking nodes to their corresponding super-nodes. Fig.~\ref{fig6}(b) shows the resulting node-level explanations after this adaptation, with importance values aggregated and visualized directly on nodes. To improve robustness, explanations were averaged over multiple non-deterministic runs, importance scores were rescaled for comparability between nodes and arcs, and salient components were selected via clustering rather than fixed thresholds, yielding a compact explanatory subgraph \cite{fiori2024gnn}. 

\subsection{Generative Explanation Methods}

Generative explanation models frame interpretability as a conditional generation task, in which explanations are synthesized rather than inferred via feature attribution \cite{chang2018counterfactual, pmlr-v119-koh20a}. In this framework, the model generates an interpretable output that captures how sensor observations give rise to human activities, enabling explanations that are aligned with human-understandable semantics. Formally, given an input sensor sequence $x$ and a predicted activity $y$, a generative explainer learns a conditional distribution $p(e \mid x,y)$, where $e$ is a human-interpretable explanatory artifact \cite{chang2018counterfactual}. Interpretability arises from the alignment between generated explanations and human-understandable activity semantics.

In \cite{sun2023gexse}, Yuan et al. introduced GeXSe (Generative Explanatory Sensor System), an intrinsically interpretable HAR framework that maps ambient sensor activations to generated visual explanations of human activities. By coupling deep generative models with sensor data, GeXSe produces human-interpretable reconstructions that illustrate how sensor events correspond to spatial and temporal activity patterns, enabling global understanding of activity routines in smart environments. In another study \cite{de2025sez}, the authors proposed SEZ-HARN, a self-explainable zero-shot HAR network that generates skeleton-based and motion-centric explanations as part of the recognition process. Their model jointly performs activity recognition and explanation generation, producing interpretable motion representations even for unseen activities. This generative formulation allows explanations to generalize beyond the training set, addressing a key limitation of post-hoc attribution methods.

\subsection{Concept-, Counterfactual-, and Prototype-Based Methods}

While attribution-based explainability dominates XAI-HAR, alternative paradigms introduce interpretability through semantic concepts, counterfactual reasoning, and prototype-based representations. These approaches move beyond feature-level relevance toward higher-level abstractions aligned with human understanding of activities and causal structure. Unlike purely post-hoc methods, they often embed interpretability directly within the model architecture or decision process.

A representative concept-based framework is X-CHAR, which introduces a concept bottleneck architecture for complex activity recognition \cite{jeyakumar2023x}. In X-CHAR, sensor streams are first mapped to interpretable sub-activities (e.g., \textit{open fridge}, \textit{pour water}), which are then temporally composed into higher-level activities. By grounding predictions in semantic concepts, explanations arise intrinsically from the model’s reasoning process. The framework also enables counterfactual analysis by examining how modifying specific concepts alters predicted outcomes, demonstrating intrinsic and counterfactual interpretability within HAR.

Counterfactual reasoning has been further explored in wearable and health-oriented HAR by AbdelRaouf et al. \cite{abdelraouf2025leveraging}, who integrated a genetic algorithm (GA)-based explanation module into a CNN–GRU–attention framework. The GA generates counterfactual instances $x_{cf}$ that minimally modify the original input $x$ to change the model prediction toward a desired class. The optimization objective combines prediction fidelity, proximity, and sparsity as

\begin{equation}
    \begin{split}
        L_{\mathrm{total}} =
        \alpha \left| f(x_{cf}) - y_{\mathrm{target}} \right|
        + \beta \sum_{i=1}^{d} (x_i - x_{cf,i})^2 \\
        + \gamma \sum_{i=1}^{d} \mathbb{I}(x_i \neq x_{cf,i}),
    \end{split}
\label{eq11}
\end{equation}

\noindent where $\alpha,\beta,\gamma \in \mathbb{R}^+$ balance prediction accuracy, $\ell_2$ proximity, and sparse feature changes, respectively. The GA searches for $x_{cf}^*$ minimizing $L_{\mathrm{total}}$ under domain constraints. Prototype-based interpretability was exemplified by DeXAR, which transformed sensor streams into structured representations and employed a prototype-constrained CNN architecture \cite{arrotta2022dexar}. In the framework, an autoencoder first mapped inputs into a latent space, where representative activity prototypes were learned. Classification was performed via similarity to these prototypes, enabling explanations in terms of resemblance to canonical activity patterns rather than abstract saliency scores. To enhance interpretability, prototype similarity was aligned with saliency-based visualizations and translated into semantic descriptions reflecting object usage, spatial context, and temporal structure.

\subsection{Hybrid and Multi-Mechanism Explainable Methods}

Beyond the primary algorithmic categories discussed above, several recent HAR frameworks adopt hybrid or multi-mechanism explainability designs that combine complementary interpretability techniques within a single system. A representative example is XTinyHAR, which investigates explainability in the context of multimodal knowledge distillation for inertial HAR \cite{lamaakal2025tiny}. XTinyHAR employs intrinsic attention rollout to analyze temporal focus within a lightweight transformer-based student model, while post-hoc IG is applied to both teacher and student networks to examine how salient motion patterns are transferred during the distillation process. This combined use of attention-based transparency and gradient-based attribution enables comparative analysis of model reasoning across architectures and training stages, providing insight into the interpretability of compressed HAR models.

To consolidate these developments, Table~\ref{tab1} summarizes the algorithmic taxonomy of XAI-HAR methods discussed in this section, highlighting the primary interpretability questions, explanation targets, and key limitations in HAR.

\begin{table*}[t]
\centering
\caption{Summary of XAI methods in HAR (Section~\ref{sec4}), highlighting the key question, explanation target, and limitations.}
\label{tab1}
\renewcommand{\arraystretch}{1.1}
\begin{tabular}{|c|p{3.0cm}|p{3.6cm}|p{2.8cm}|p{3.8cm}|}
\hline
\textbf{Section} & \textbf{XAI Method Category} & \textbf{Primary HAR Question Answered} & \textbf{Explanation Target} & \textbf{Key Limitation in HAR Context} \\
\hline

IV-A &
Perturbation \& Sampling (LIME, SHAP, Anchors, PFI) &
Which input features or temporal segments influenced this prediction? &
Local attribution; some global importance &
Weak temporal consistency; sensitive to correlated features \\
\hline

IV-B &
Gradient \& Path-Based (Saliency, IG, Grad-CAM) &
Which timesteps or channels were most influential? &
Local, model-internal (input/activation space) &
Correlation-based, not causal; sensitive to gradients and baselines \\
\hline

IV-C &
Surrogate \& Symbolic Methods &
How can this decision be approximated in an interpretable form? &
Rules, trees, symbolic/temporal logic &
Lose fine-grained temporal dynamics and sequential dependencies \\
\hline

IV-D &
Graph-Structured Methods &
How do sensors or entities interact structurally? &
Nodes, edges, subgraphs (spatio-temporal) &
Limited empirical validation and scalability in HAR \\
\hline

IV-E &
Generative Methods &
How can sensor data generate semantic explanations of activities? &
Generated semantic or visual artifacts &
High complexity; difficult to ver-
ify faithfulness to underlying
model reasoning \\
\hline

IV-F &
Concept / Counterfactual / Prototype &
What concepts, counterfactual alternatives, or prototypes define this activity? &
Concepts, counterfactuals, prototype similarity &
Limited coverage; weak temporal consistency \\
\hline

IV-G &
Hybrid Methods &
Can multiple explanation mechanisms be combined effectively? &
Feature-, temporal-, and semantic-level explanations &
Increased complexity; lack of standardized evaluation \\
\hline

\end{tabular}
\end{table*}

\section{Technological Summary and Discussions}
\label{sec5}

\subsection{Technological Synthesis}

The studies reviewed in sections~\ref{sec3} and \ref{sec4} reveal that XAI-HAR methods differ not only in their computational mechanisms but also in the types of interpretability they provide and the trade-offs they impose for real-world systems. Perturbation- and sampling-based methods remain among the most flexible approaches because they can be applied to black-box HAR models without access to model internals. In practice, LIME has been effective for sample-level debugging and post-hoc inspection in wearable, EEG, and smart-home settings \cite{uddin2021human,hussain2023explainable,liu2024enhanced,das2023explainable}. However, its effectiveness depends strongly on perturbation strategy, feature representation, and temporal segmentation, making it sensitive to correlated sensor streams and long-duration activities \cite{hussain2023explainable,bijalwan2024interpretable,das2023explainable}. By contrast, SHAP supports both local and global analysis and often provides more stable and globally coherent attributions, particularly in long-horizon and smart-home settings \cite{liu2024enhanced,benos2025explainable,bijalwan2024interpretable,das2023explainable,tempel2025explaining}, although its computational cost can increase with model complexity \cite{liu2024enhanced,das2023explainable}. Anchors further complement these methods by offering high-precision, rule-based local explanations in ambient HAR, albeit with limited coverage across instances \cite{das2023explainable}, while permutation-based attribution supports global interpretability by revealing dominant sensing modalities and enabling model simplification \cite{tokas2025lightweight}.

Gradient- and path-based methods provide a different technological advantage by directly exploiting model derivatives and internal representations. Saliency methods are computationally efficient and have been applied across accelerometer, audio, and radar-based HAR, but their explanations are often unstable because of noise and gradient saturation \cite{aquino2023explaining,kim2025granular,waghumbare2025diat}. IG improves on this by producing smoother and more temporally consistent local attributions for deep temporal models \cite{lamaakal2025tiny}. However, these methods often reflect sensitivity rather than true causal importance and may suffer from instability or dependence on baseline selection. Grad-CAM addresses some of these limitations by offering architecture-aware explanations that are tightly coupled to convolutional representations and can highlight discriminative temporal regions and dominant channels more coherently than perturbation methods \cite{mekruksavanich2025efficient,huang2024explainable}. Yet its reliance on intermediate convolutional features limits explicit feature-level attribution, making it most effective as a complement to model-agnostic approaches such as LIME and SHAP \cite{pellano2024movements,mekruksavanich2025efficient}.

Surrogate and symbolic approximation methods shift the focus from feature relevance to decision logic. Rule-based and tree-based surrogates provide transparent, logic-driven explanations that are especially attractive in safety-critical, healthcare, and smart-home HAR applications \cite{bettini2021explainable}. Their strengths lie in auditability, semantic clarity, and ease of validation, but this interpretability often comes at the cost of expressive power and fidelity to complex temporal dynamics \cite{bettini2021explainable}. Logic-induction and formal reasoning approaches extend this paradigm by producing temporally grounded symbolic explanations, though they generally require structured representations and higher modeling complexity \cite{cao2023discovering,l2021probabilistic}.

Graph-structured explanation methods are particularly valuable when relational structure is central to activity semantics. By operating on nodes, edges, and subgraphs rather than flat feature vectors, they preserve spatial and temporal dependencies among sensors and activity components \cite{ying2019gnnexplainer}. In HAR, GNN-XAR shows that graph-based explainability can align explanations with relational sensor representations and translate them into semantically meaningful descriptions of sensor interactions \cite{fiori2024gnn}. This makes graph methods especially suitable for smart-home and other structured HAR settings, although their empirical use remains more limited than attribution-based approaches \cite{fiori2024gnn}.

Generative methods, in contrast, move interpretability beyond attribution by synthesizing semantic explanatory artifacts. GeXSe and SEZ-HARN show that explanations can be generated as visual, motion-centric, or activity-level reconstructions that align more naturally with human understanding \cite{sun2023gexse,de2025sez}. Their strength lies in shifting explainability from diagnostic feature scoring toward semantic reconstruction and human-centered communication, but this comes with increased modeling complexity and added difficulty in verifying faithfulness to the underlying recognition process \cite{sun2023gexse,de2025sez}.

Concept-, counterfactual-, and prototype-based methods further extend interpretability by grounding explanations in human-understandable abstractions such as activity components, alternative scenarios, or representative patterns. X-CHAR shows how concept bottlenecks can make complex activity recognition intrinsically interpretable by mapping sensor streams to sub-activities before composing higher-level activities \cite{jeyakumar2023x}. Counterfactual frameworks add actionable interpretability by identifying minimal changes that alter predictions, while prototype-based methods explain decisions through similarity to canonical activity patterns \cite{abdelraouf2025leveraging,arrotta2022dexar}. Collectively, these paradigms align model reasoning more closely with human-understandable activity semantics, although they remain less prevalent and less extensively validated than attribution-based approaches \cite{jeyakumar2023x,abdelraouf2025leveraging,arrotta2022dexar}.

Hybrid and multi-mechanism methods suggest that no single XAI paradigm is sufficient for all HAR scenarios. Frameworks such as XTinyHAR combine attention rollout with Integrated Gradients to analyze both temporal focus and transferred motion saliency in compressed multimodal HAR systems \cite{lamaakal2025tiny}. Their main advantage is support for multi-level interpretability across feature, temporal, and semantic perspectives, but this comes with increased methodological complexity and a lack of unified evaluation criteria. Overall, the technological trajectory of XAI-HAR is moving away from isolated explanation tools toward integrated, mechanism-aware interpretability frameworks tailored to the temporal, multimodal, and semantic demands of HAR.

These findings suggest that XAI-HAR is evolving from isolated explanation techniques toward integrated, system-level interpretability frameworks that must balance fidelity, usability, and computational efficiency. Importantly, the effectiveness of an explanation method is not solely determined by its theoretical properties, but by how well it supports human understanding and decision-making in real-world environments.

\subsection{Open Challenges}

Despite significant progress, several challenges remain that limit the reliability and practical adoption of XAI-HAR systems.

1) \textit{Evaluation of explanation quality:}
A major challenge is the lack of standardized evaluation protocols tailored to HAR. Although faithfulness, stability, and consistency are increasingly discussed, current studies rely on heterogeneous evaluation practices, making cross-method comparison difficult and weakening the reliability of interpretability claims.

2) \textit{Temporal coherence and sequential dependencies:}
HAR is inherently sequential, with activities unfolding over long durations and evolving phases. Many existing methods remain dominated by local feature relevance and do not adequately capture temporal continuity, activity transitions, or long-range dependencies.

3) \textit{Semantic gap between sensor data and activity meaning:}
Explanations are often grounded in low-level sensor features such as channels, windows, or latent activations, whereas human understanding requires higher-level semantic concepts. Bridging this gap is essential for user trust, particularly in healthcare and assistive applications.

4) \textit{Multimodality and relational structure:}
Modern HAR systems rely on multiple sensing modalities and structured interactions across sensors and entities. However, explainability methods are still more mature for flat feature representations than for multimodal and relational settings, limiting their ability to capture cross-sensor dependencies.

5) \textit{Robust and deployable XAI-HAR systems:}
Real-world deployment requires balancing interpretability, fidelity, robustness, and computational efficiency under noisy and subject-dependent conditions. Moving beyond isolated demonstrations toward reliable, scalable, and deployable XAI-HAR frameworks remains a key challenge.

\section{Conclusions}
\label{sec6}

This study presents a comprehensive review of XAI in HAR, integrating sensing modalities, conceptual frameworks, and algorithmic taxonomies into a unified perspective. The analysis shows that explainability in HAR is inherently multi-dimensional, requiring approaches that capture feature-level attribution, temporal dynamics, and semantic interpretation. It further highlights that no single explanation paradigm is sufficient, and that effective XAI-HAR systems rely on combining complementary mechanisms to address the challenges of sequential, multimodal, and noisy sensor data. From a system perspective, the effectiveness of explainability depends not only on algorithmic design but also on how well explanations support human understanding and decision-making in real-world applications, particularly in domains such as healthcare and assistive systems.

Future work should focus on bridging the gap between low-level sensor representations and human-understandable activity semantics, improving temporal coherence in explanations, and advancing methods that capture multimodal and relational dependencies. In parallel, the development of standardized evaluation frameworks remains essential for assessing explanation quality in a consistent and reliable manner. Progress in XAI-HAR therefore depends on developing explanation methods that are not only theoretically grounded but also interpretable, reliable, and practical for real-world deployment.

\bibliographystyle{IEEEtran}
\bibliography{reference.bib}

\end{document}